\documentclass[preprint,12pt]{elsarticle}

\usepackage{amssymb}
\usepackage{amsmath}
\usepackage{booktabs}
\usepackage{graphicx}
\usepackage{xurl}
\usepackage{lineno}
\usepackage[section]{placeins}  % auto FloatBarrier before each \section

\journal{Science of Remote Sensing}

\begin{document}

\begin{frontmatter}

\title{Morphology-Guided Cross-Task Coupling for Joint Building Height and Footprint Estimation}

\author[skku]{Jinzhen Han}
\author[skku]{JinByeong Lee}
\author[leeds]{Jisung Kim\corref{cor1}}
\ead{gyjki@leeds.ac.uk}
\author[skku]{HongSik Yun\corref{cor1}}
\ead{yoonhs@skku.edu}
\cortext[cor1]{Corresponding authors.}

\affiliation[skku]{organization={Department of Civil, Architectural and Environmental System Engineering, Sungkyunkwan University},
            addressline={2066 Seobu-ro, Jangan-gu},
            city={Suwon},
            postcode={16419},
            country={Republic of Korea}}

\affiliation[leeds]{organization={School of Geography, University of Leeds},
            city={Leeds},
            postcode={LS2 9JT},
            country={United Kingdom}}

%% ---- Abstract ----
\begin{abstract}
Building height (BH) and building footprint (BF) jointly describe the vertical and horizontal extent of the built environment and are required inputs for urban climate, disaster-risk, and population-mapping models. The two parameters are coupled through floor-area-ratio (FAR) constraints, yet remote-sensing approaches typically treat them as independent regression targets. We argue that explicitly encoding this cross-task coupling is more impactful than further refining individual encoders, and propose \textsf{MorphoFormer}, a joint BH/BF estimation framework built around two complementary mechanisms: (i) a BF-Guided Task Decoder (BGTD) that gates the height branch via cross-attention on a footprint-derived morphology context, and (ii) a Morphology Consistency Loss (MCL) that supervises a height-from-footprint surrogate against the ground-truth BH, indirectly forcing the BF feature to encode height-correlated structure. The encoder is a single-stage Swin backbone fed by Sentinel-1 SAR, Sentinel-2 multispectral, and DEM inputs, trained and evaluated on a geo-blocked split of 54 cities. Against a Swin-MTL baseline at identical receptive field, MorphoFormer reduces BH test RMSE from 3.39 to 3.15\,m ($R^{2}$ improves 0.62 $\to$ 0.67) with BF $R^{2}$ stable at 0.80. Controlled ablations at identical capacity attribute most of this 0.24\,m improvement to the two proposed mechanisms: removing BGTD raises BH RMSE by 0.11\,m and removing MCL raises it by 0.11\,m, with the residual $\approx 0.02$\,m falling within the noise floor of encoder-side variations. Because both mechanisms act on cross-task representations rather than pixels, the design carries no intrinsic dependence on input resolution.
\end{abstract}

%% ---- Highlights ----
\begin{highlights}
\item BH and BF are FAR-coupled; we encode this prior at feature and output levels.
\item BGTD: a BF-guided cross-gate routes morphology context into the height branch.
\item MCL: an auxiliary loss aligns a height-from-footprint surrogate with true BH.
\item BGTD and MCL jointly account for the bulk of the test-set BH improvement.
\item BH RMSE 3.15 vs.\ 3.39\,m at identical 9$\times$9 RF on a 54-city geo-blocked split.
\end{highlights}

%% ---- Keywords ----
\begin{keyword}
building height \sep building footprint \sep deep learning \sep
multi-task learning \sep cross-task coupling \sep morphology consistency \sep
Sentinel-1 \sep Sentinel-2
\end{keyword}

\end{frontmatter}

% \linenumbers  % disabled for arXiv preprint build; re-enable for journal review submission

%% =============================================================
%% 1. Introduction (TBD)
%% =============================================================
\section{Introduction}
\label{sec:intro}

In recent decades, urban areas worldwide --- and Asian megacities in particular --- have undergone pronounced three-dimensional transformation, marked not only by horizontal expansion but also by accelerating vertical densification \cite{fang2016changing, frolking2013global, frolking2024global}. Quantifying this transformation requires two morphological parameters in tandem: building height (BH), which captures the vertical extent of the built environment, and building footprint (BF), which captures its horizontal occupancy. Together these parameters are required inputs for urban climate and heat-island modelling \cite{xi2021impacts, perini2014effects}, flood-exposure assessment \cite{huang2020estimates}, cascading-disaster simulation including post-earthquake fire propagation \cite{tian2025fire}, urban energy modelling \cite{reinhart2016urban}, and population mapping \cite{tatem2017worldpop, schiavina2023ghspop}. Yet globally consistent BH/BF data remain unevenly available: even in data-rich regions, fewer than half of buildings include height attributes, and large parts of the Global South lack systematic 3-D urban records altogether \cite{geofabrik_osm_2018}.

A long line of work has sought to fill this gap from remote sensing. The highest per-building accuracies are obtained from very-high-resolution (VHR) optical imagery \cite{buyukdemircioglu2022deep, li2021deep, rastogi2022automatic}, airborne LiDAR \cite{park2019creating, li2020developing}, and VHR SAR \cite{sun2019large}, all of which can deliver root-mean-squared errors below 2\,m on individual buildings but rely on data sources whose cost, footprint, and revisit prevent global, repeatable deployment \cite{cai2024automated}. Open-access Sentinel-1 SAR and Sentinel-2 multispectral imagery have therefore become the dominant medium for grid-based, scalable BH/BF retrieval \cite{frantz2021national, wu2023first, cai2023deep, chen2024refining}. Recent work has further pushed coverage and accuracy through hybrid CNN--Transformer encoders \cite{wang2024mfbhnet}, ICESat-2--Sentinel fusion \cite{zheng2025nectnet}, OpenStreetMap-conditioned per-building reconstruction \cite{mostafavi2024utglobus}, and commercial-imagery-driven 3\,m maps \cite{zhu2025globalbuildingatlas}. These advances are real, but a recurring trade-off persists: each step toward finer resolution or broader coverage is paid for in proprietary data, ancillary vector layers, or restricted geographic validation.

Beyond this data-side trade-off lies a model-side gap that persists regardless of input resolution. Across nearly all of the literature, BH and BF are treated as independent regression targets, even though they are tightly coupled at the morphological level: the two parameters jointly trace the floor-area-ratio (FAR), and the FAR distribution is in turn structured by zoning regimes and built-form typology codified in local-climate-zone taxonomies \cite{stewart2012local, ching2018wudapt}. For retrieval purposes, however, this coupling is almost always left to the data to discover. Multi-task formulations are common, but in their predominant form --- a shared encoder feeding two parallel regression heads \cite{shafts2023} --- they share only \emph{representation} between the tasks and impose no \emph{structural} constraint at the morphological level. The FAR prior is present in the labels but absent in the architecture and absent in the loss; whether the input is 100\,m Sentinel-derived or sub-meter VHR, this gap remains the same.

\begin{figure*}[!htbp]
\centering
\includegraphics[width=\textwidth]{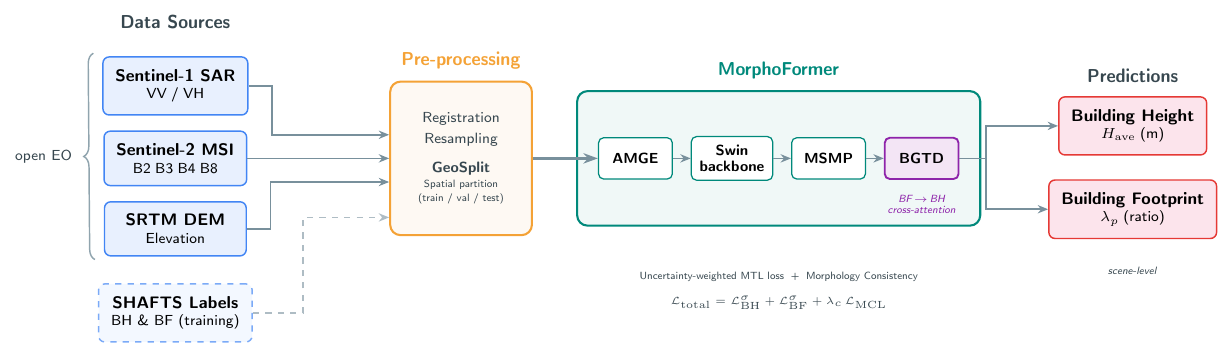}
\caption{Overview of the proposed \textsf{MorphoFormer} framework. The BGTD module (highlighted) and the Morphology Consistency Loss ($\mathcal{L}_{\mathrm{MCL}}$) jointly operationalize the cross-task coupling described in Section~\ref{sec:method}.}
\label{fig:workflow}
\end{figure*}

We argue that operationalizing this prior requires changes at \emph{both} the representation level and the output level, and propose \textsf{MorphoFormer}, a joint BH/BF estimation framework whose two core ingredients act at these two levels respectively. The first is a BF-Guided Task Decoder (BGTD): the footprint branch encodes a compact morphology context, and the height branch attends to this context through cross-attention before regressing height. The second is a Morphology Consistency Loss (MCL): an auxiliary head reads height directly from the footprint feature, and the loss penalizes its disagreement with ground-truth height. Both mechanisms act on cross-task representations rather than on pixel arrays. We instantiate them on a single-stage Swin backbone fed by Sentinel-1 SAR, Sentinel-2 multispectral, and DEM inputs, and validate on a geo-blocked split of 54 cities (Fig.~\ref{fig:workflow}). We use the open Sentinel/DEM stack at 100\,m as a reproducible experimental medium for the proposed mechanism rather than as an end in itself: nothing in BGTD or MCL ties them to a specific input resolution, and the same routing principle is expected to apply directly to 10\,m ARD or sub-meter VHR inputs --- the contribution of this work is the cross-task mechanism, not a particular data product.

The contributions of this work are:
\begin{enumerate}
    \item We identify the explicit encoding of the FAR-induced (BH,\,BF) coupling as an under-exploited direction for joint retrieval, and articulate the gap left by shared-encoder multi-task formulations.
    \item We propose \textsf{MorphoFormer}, which operationalizes this prior through two mechanisms at distinct levels: BGTD (feature-level cross-attention gating) and MCL (output-level consistency, in which a height-from-footprint surrogate is supervised against the ground-truth BH).
    \item Against a Swin-MTL baseline at identical receptive field, MorphoFormer reduces BH RMSE from 3.39 to 3.15\,m on a 54-city geo-blocked test split.
    \item Controlled ablations at identical capacity attribute most of the 0.24\,m gap to BGTD ($+0.11$\,m if removed) and MCL ($+0.11$\,m if removed), with the encoder pipeline (AMGE, MSMP) contributing within encoder-side noise.
\end{enumerate}

%% =============================================================
%% 2. Data and Pre-processing
%% =============================================================
\section{Data and Pre-processing}
\label{sec:data}

\subsection{Reference Data}
\label{sec:data:ref}

We use the open-source SHAFTS Reference Dataset (v2022.3) \cite{shafts2023}, which discretizes vector building inventories from authoritative cadastres and OpenStreetMap onto a regular 100\,m\,$\times$\,100\,m grid via Fishnet Analysis \cite{musiaka2021}, and validates the resulting grid-cell labels against LiDAR-based digital surface models. Each cell carries a footprint ratio $\lambda_{p}$ and an area-weighted average building height $H_{\mathrm{ave}}$,
\begin{align}
\lambda_{p}      &= \frac{\sum_{i \in \mathcal{J}} A_{i}}{(100\,\mathrm{m})^{2}}, \\
H_{\mathrm{ave}} &= \frac{\sum_{i \in \mathcal{J}} A_{i}\,h_{i}}{\sum_{i \in \mathcal{J}} A_{i}},
\end{align}
where $\mathcal{J}$ is the set of buildings intersecting the cell, $A_{i}$ is the intersection area of the $i$th building with the cell, and $h_{i}$ is its measured height. From SHAFTS we retain 54 cities for which both labels are available, and use the cell-level $(\lambda_{p}, H_{\mathrm{ave}})$ as scene-level supervision targets.

\subsection{Explanatory Data}
\label{sec:data:exp}

Three openly accessible sources are used as explanatory inputs (Table~\ref{tab:data_sources}): Sentinel-1 SAR provides scattering-based vertical-structure cues, Sentinel-2 multispectral imagery provides material and texture cues, and SRTM DEM provides a terrain prior that disentangles relief from building height. All three are retrieved per city through an automated Google Earth Engine pipeline that performs cloud filtering, annual percentile aggregation, and city-extent clipping, yielding temporally consistent and spatially aligned stacks across all 54 cities.

\begin{table}[!htbp]
  \centering
  \small
  \caption{Explanatory data sources and specifications.}
  \label{tab:data_sources}
  \begin{tabular}{p{0.32\linewidth} p{0.55\linewidth}}
    \toprule
    \textbf{Source} & \textbf{Specification} \\
    \midrule
    Sentinel-1 GRD Level-1 & 10\,m, VV and VH polarizations \\
    Sentinel-2 Level-2A    & 10\,m, blue, green, red, near-infrared (B2, B3, B4, B8) \\
    SRTM V3 DEM            & 30\,m elevation, resampled to 10\,m \\
    \bottomrule
  \end{tabular}
\end{table}

\subsection{Sample Construction and Filtering}
\label{sec:data:sample}

\begin{figure*}[!htbp]
  \centering
  \includegraphics[width=0.95\textwidth]{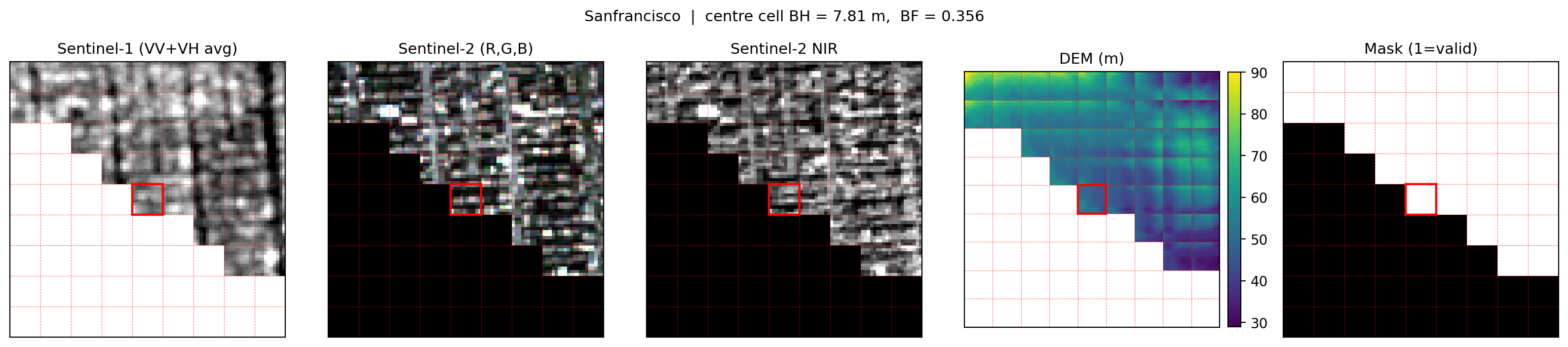}
  \caption{A $90\times90$ input scene from the test split (San Francisco, coastline). Panels left to right: Sentinel-1 VV/VH average, Sentinel-2 RGB, Sentinel-2 NIR, SRTM DEM, and the validity mask. Dashed red lines mark cell boundaries; the regression-target centre cell is outlined in solid red.}
  \label{fig:input_tensor}
\end{figure*}

Each sample is an input scene paired with one scalar pair $(\lambda_{p}, H_{\mathrm{ave}})$. The scene is a $9 \times 9$ grid window centred on the target cell with each cell rendered as a $10 \times 10$ pixel block, giving a $90 \times 90$ pixel tensor across the eight explanatory bands plus a validity mask (Fig.~\ref{fig:input_tensor}); the label is taken as the centre cell's $(\lambda_{p}, H_{\mathrm{ave}})$, with the eight neighbours serving as spatial context. The validity mask is part of the input rather than a post-hoc weight: cells that fall on water or outside the urban extent carry no SHAFTS reference and would otherwise be indistinguishable from low-rise built cells in the band signals alone, so the mask channel makes their invalidity explicit to the encoder. We discard samples that violate any of three plausibility rules: $H_{\mathrm{ave}} \notin [2,\,500]$\,m, $\lambda_{p} \le 0.01$, or $\lambda_{p} < 0.04$ together with $H_{\mathrm{ave}} \ge 20$\,m (sliver-with-tall-structure artefacts).

\subsection{Spatially Stratified Split}
\label{sec:data:split}

Because the explanatory cache is per-band rather than per-window, the $9 \times 9$ context is assembled at runtime from the eight neighbours of each centre cell. Under a naively random split, an expected $\sim$80\,\% of any test cell's 80 neighbours are themselves training cells whose pixels were already seen during training (as centres of their own training samples), so the same raw band values appear in train and test inputs and the model is implicitly evaluated on partially memorized terrain. This concern is sharpened by the analysis of Section~\ref{sec:prior}: the (BH,\,BF) coupling we target is context-conditional with strong city-to-city heterogeneity, so a split that mixes neighbouring cells across subsets would not only inflate accuracy through pixel-level leakage but also prevent any test-time gain from being cleanly attributed to a learned cross-task mechanism rather than to memorized local context.

\begin{figure*}[!htbp]
  \centering
  \includegraphics[width=\textwidth]{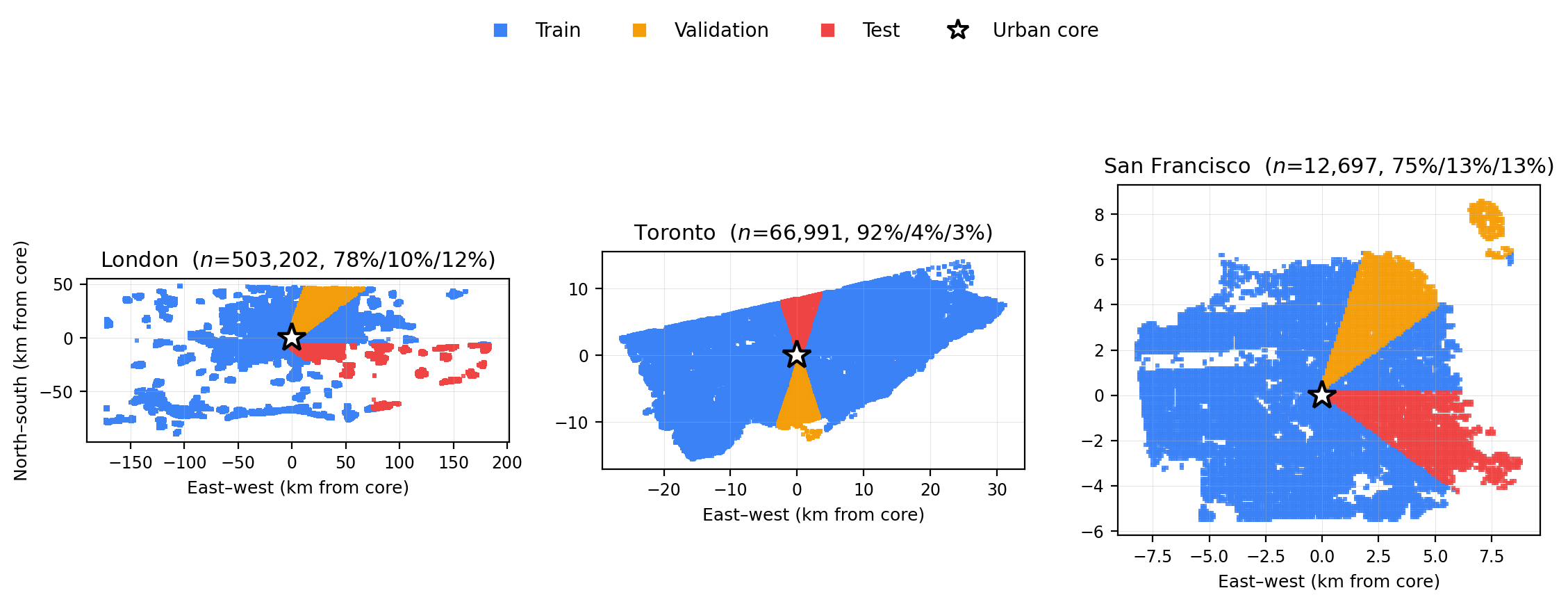}
  \caption{GeoSplit assignment for three cities of contrasting morphology and extent. Each marker is a 100\,m cell coloured by its train / validation / test assignment; the urban core is marked with a star.}
  \label{fig:geosplit}
\end{figure*}

We therefore adopt a city-internal, geographically stratified partition (\emph{GeoSplit}) at an 8\,:\,1\,:\,1 ratio. Each city is divided into ten equal-angle radial sectors centred on the urban core, and the sectors are assigned to the three subsets by a greedy balancer that approximates the target ratio while keeping both central and peripheral sectors in every subset (Fig.~\ref{fig:geosplit}). Each sector is a contiguous wedge, so most of its interior cells have all eight neighbours inside the same sector and leakage is confined to a thin layer at sector boundaries. The radial wedge geometry is deliberate: the dominant within-city non-stationarity in built form is the radial gradient from dense urban core to sparse periphery, and a wedge cuts \emph{across} this gradient rather than along it, so every subset retains both core and peripheral regimes of every retained city.

%% =============================================================
%% 3. The Cross-Task Coupling Prior
%% =============================================================
\section{The Cross-Task Coupling Prior}
\label{sec:prior}

A premise of this work is that BH and BF are not statistically independent, and that any architecture which treats them as such leaves predictive information on the table. Before specifying any mechanism, we therefore characterize the strength and structure of this coupling on the training labels alone, independently of any model.

\begin{figure*}[!htbp]
  \centering
  \includegraphics[width=0.95\textwidth]{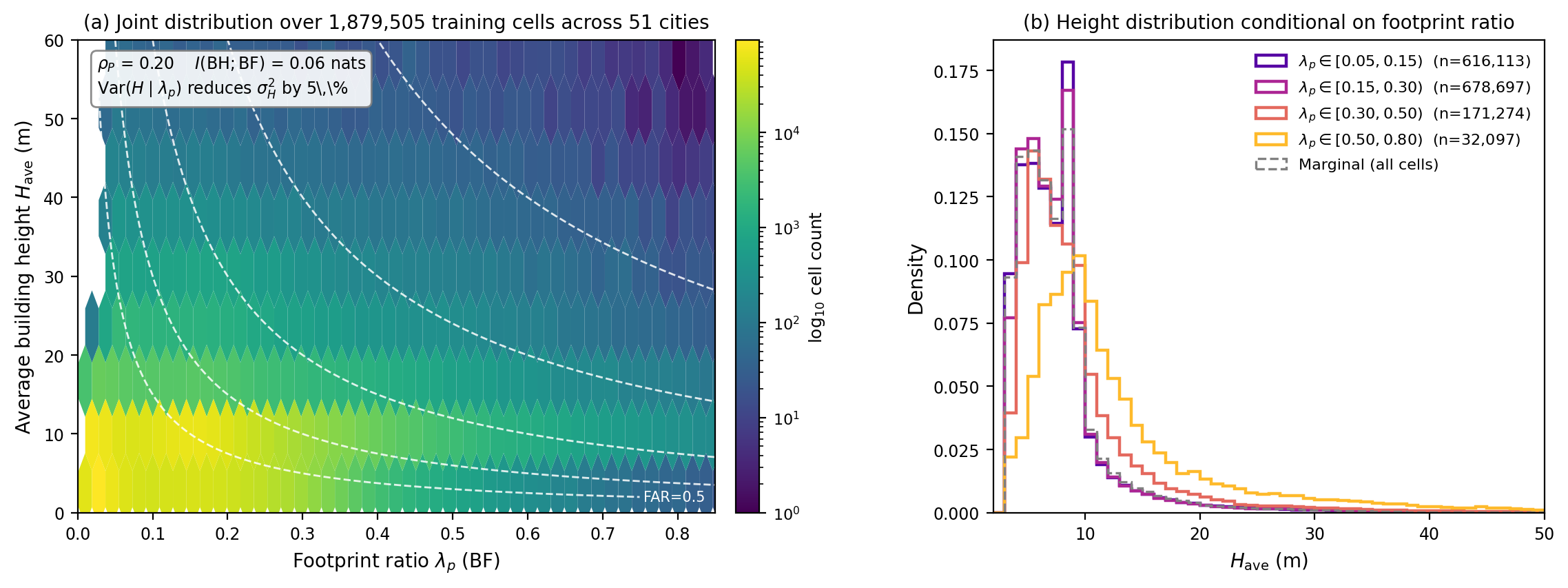}
  \caption{The (BH,\,BF) coupling on the training split. (a) Joint hexbin of $H_{\mathrm{ave}}$ and $\lambda_{p}$ over 1.88\,M cells, with FAR iso-curves overlaid. (b) Conditional $H_{\mathrm{ave}}$ distributions by $\lambda_{p}$ bin.}
  \label{fig:far_prior}
\end{figure*}

Figure~\ref{fig:far_prior}(a) shows the joint distribution of $H_{\mathrm{ave}}$ and $\lambda_{p}$ over the 1.88\,M retained training cells. The two parameters are jointly bounded by floor-area-ratio physics, and we overlay iso-curves of $\mathrm{FAR} = \lambda_{p}\,H_{\mathrm{ave}}/h_{\mathrm{floor}}$ with $h_{\mathrm{floor}} = 3$\,m: cells concentrate within FAR\,$\lesssim\,4$ in the residential and mid-rise regime, with a sparse upper-right tail populated by dense high-rise. The \emph{marginal} coupling captured by this joint distribution is real and stable across the training set: Pearson $\rho_{P} = 0.20$, Spearman $\rho_{S} = 0.16$, with 4.6\,\% of $\mathrm{Var}(H_{\mathrm{ave}})$ explained by $\lambda_{p}$ alone. This places a tight, data-driven ceiling on what a marginal-only predictor can extract --- a 4.6\,\% variance reduction is at most a 2.3\,\% RMSE reduction, or $\approx 0.08$\,m on the 3.39\,m baseline --- and this ceiling is the threshold our mechanism is designed to clear by routing cross-task signal that the marginal distribution alone cannot expose. Panel (b) makes the structure visible: the conditional $H_{\mathrm{ave}}$ distributions shift right monotonically as $\lambda_{p}$ increases, with substantial within-bin spread, so a global function $H = f(\lambda_{p})$ captures the conditional mean while leaving the bulk of the conditional variance available for cross-task mechanisms to recover.

What this picture rules out is that any cross-task structure beyond the $\approx 0.08$\,m ceiling will be available as a function of $\lambda_{p}$ alone. A multi-task model that merely shares an encoder between BH and BF heads is therefore \emph{implicitly bounded by precisely this ceiling}: it captures only what the marginal coupling exposes, with no explicit pathway by which BF-derived information is required to inform the BH prediction. The shared-encoder formulation that dominates the BH/BF retrieval literature \cite{shafts2023} sits in this regime --- a joint representation is allowed to form, but no architectural commitment requires it to. This is the gap our architecture targets: rather than rely on the encoder to surface cross-task structure as a side effect of joint training, we route the BF-derived signal into the BH prediction along two explicit pathways. Section~\ref{sec:method} introduces both: a feature-level cross-attention gate (BGTD, Section~\ref{sec:method:bgtd}) and an output-level consistency penalty (MCL, Section~\ref{sec:method:mcl}); Section~\ref{sec:results} then reports whether either mechanism in fact clears the $\approx 0.08$\,m ceiling that the marginal coupling sets.

%% =============================================================
%% 4. Method
%% =============================================================
\section{Method}
\label{sec:method}

\begin{figure*}[!htbp]
  \centering
  \includegraphics[width=\textwidth]{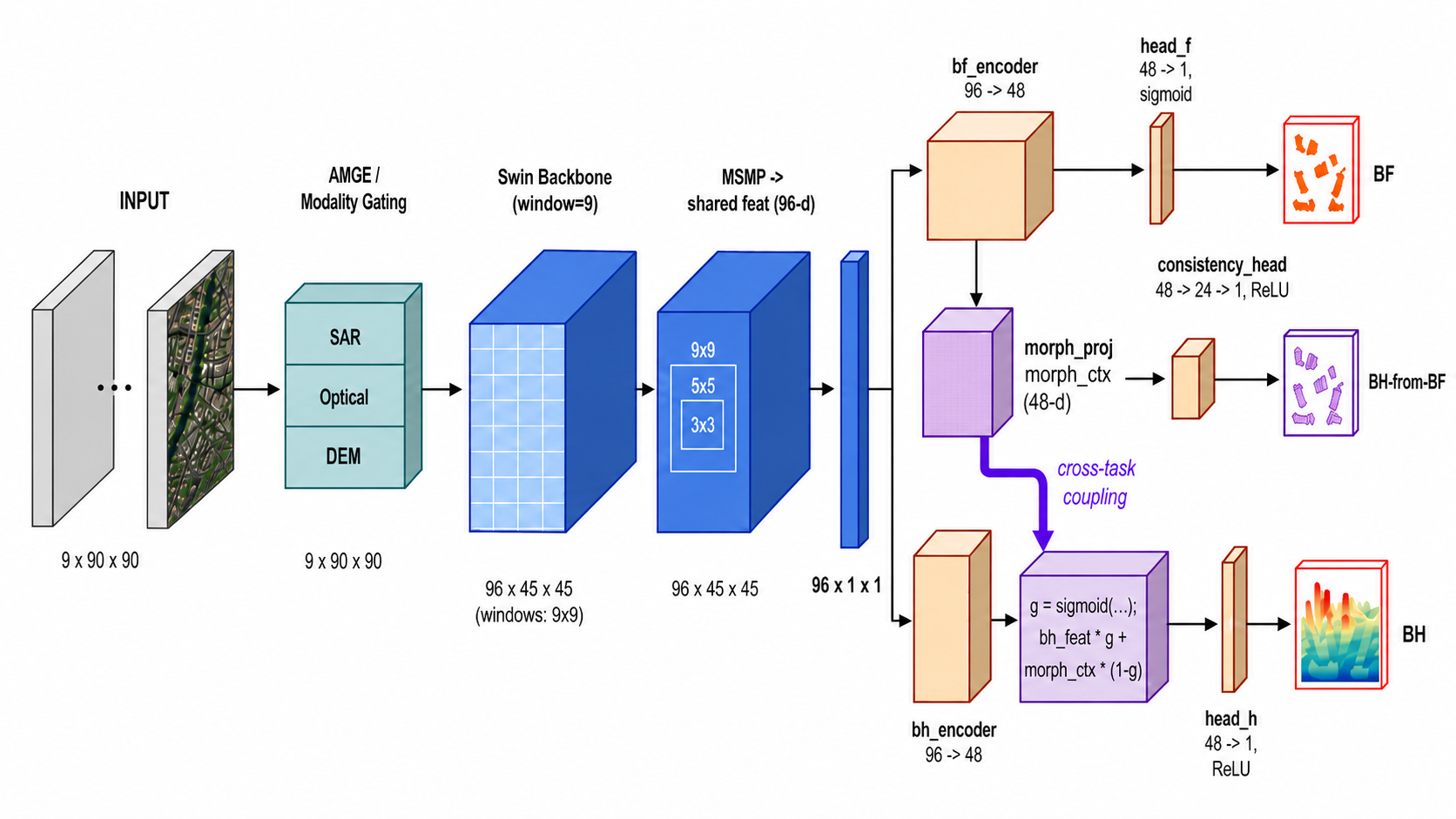}
  \caption{Architecture of MorphoFormer. Encoder pipeline (Section~\ref{sec:method:backbone}) on the left; BGTD (Section~\ref{sec:method:bgtd}) and the auxiliary head producing $\widehat{\mathrm{BH}}_{\mathrm{from\,BF}}$ for MCL (Section~\ref{sec:method:mcl}) on the right. Cross-task pathways highlighted in purple.}
  \label{fig:architecture}
\end{figure*}

The MorphoFormer architecture (Fig.~\ref{fig:architecture}) is organized around the prior identified in Section~\ref{sec:prior}: the joint (BH,\,BF) distribution carries structure that an independent two-headed regressor cannot exploit. We operationalize this structure through two complementary mechanisms acting at different levels of the network: a feature-level coupling realized by the BF-Guided Task Decoder (Section~\ref{sec:method:bgtd}), and an output-level coupling realized by the Morphology Consistency Loss (Section~\ref{sec:method:mcl}). Section~\ref{sec:method:backbone} describes the encoder pipeline that feeds both mechanisms.

\subsection{Backbone: Modality-Aware Encoding and Multi-Scale Pooling}
\label{sec:method:backbone}

The encoder pipeline converts the nine-channel input scene into a single shared feature vector that both task branches consume. It is composed of three components, described briefly below.

\textbf{AMGE} (Adaptive Modality Gating Encoder) applies a squeeze-and-excitation-style gate independently to each modality group --- SAR (2 bands), optical (4 bands), and DEM (1 band) --- before they are concatenated. Each gate performs global average pooling, a two-layer MLP with sigmoid output, and a channel-wise rescaling, learning a per-band confidence weight conditioned on the scene as a whole. The validity mask is passed through unmodified. The role of AMGE is calibration: it lets the network down-weight, for example, optical bands in cloudy scenes or DEM in flat regions, before any cross-modality mixing occurs.

\textbf{Swin backbone.} The gated input is then encoded by a single-stage Swin Transformer variant \cite{liu2021swin}. We use $2\times2$ patch embedding (yielding $45\times45$ tokens at $D_{0}=96$ channels), two SwinBlocks with shifted-window self-attention at window size $9\times9$, and three attention heads. Restricting the backbone to a single stage and a small embedding dimension keeps the total parameter count of the encoder pipeline below $0.4$\,M, which keeps the contribution comparison of Section~\ref{sec:results} unambiguous: any accuracy gain attributed to BGTD or MCL is not an artefact of capacity differences.

\textbf{MSMP} (Multi-Scale Morphology Pooling) collapses the encoded token map into the shared feature vector by aggregating three centre crops at $3\times3$, $5\times5$, and $9\times9$ tokens around the target cell. Each crop is reduced to a $D_{0}$-dim vector by global average pooling, and the three vectors are fused by a softmax-attention operator that learns a sample-conditional weighting over the three scales. The output is a single 96-dimensional shared feature $\mathbf{h}$, which is the entry point of the BGTD decoder of Section~\ref{sec:method:bgtd}.

The 9-cell window aggregated by this pipeline corresponds to a 900\,m receptive field on the 100\,m label grid, which is wider than the per-cell or 3-cell contexts typical of grid-based BH/BF retrieval. This gives the encoder access to neighbourhood-scale morphological transitions --- density gradients, block-to-block discontinuities, and waterfront/industrial boundaries --- that single-cell formulations cannot resolve, and is the substrate on which the cross-task mechanisms of Sections~\ref{sec:method:bgtd}--\ref{sec:method:mcl} operate.

\subsection{BGTD: Feature-Level Cross-Task Coupling}
\label{sec:method:bgtd}

\paragraph{Motivation.} Section~\ref{sec:prior} established that the marginal $(\lambda_{p}, H_{\mathrm{ave}})$ coupling sets a tight $\approx 0.08$\,m ceiling on what a shared-encoder predictor can extract from $\lambda_{p}$ alone, and that exploiting cross-task structure beyond this ceiling requires an explicit BF-to-BH routing pathway that a shared encoder feeding two parallel regression heads does not commit to. BGTD introduces such a pathway directly into the decoder: the BH branch reads not only the shared representation but also a compact morphology context distilled from the BF branch, and a learned per-channel gate decides how much of each to admit.

\paragraph{Architecture.} Let $\mathbf{h}\in\mathbb{R}^{D_{0}}$ be the shared feature vector emerging from MSMP, with $D_{0}=96$ in our default configuration, and let $D=D_{0}/2=48$ be the per-task hidden dimension. The BF branch first encodes a task-specific feature
\[
\mathbf{b}_{f} = \mathrm{GELU}\bigl(\mathrm{LN}(W_{bf}\mathbf{h})\bigr) \in \mathbb{R}^{D},
\]
from which the regressed footprint ratio is read off by a sigmoid head $\hat{\lambda}_{p} = \sigma(W_{f}\mathbf{b}_{f})$. The same feature is then projected into a morphology context vector
\[
\mathbf{m} = \mathrm{GELU}\bigl(\mathrm{LN}(W_{m}\mathbf{b}_{f})\bigr) \in \mathbb{R}^{D},
\]
which is the only signal the BH branch is allowed to import from the BF lane. The BH branch independently encodes its own feature $\mathbf{b}_{h} = \mathrm{GELU}(\mathrm{LN}(W_{bh}\mathbf{h}))$, and BGTD then computes a per-channel gate
\begin{equation}
\mathbf{g} \;=\; \sigma\bigl( W_{g}\,[\,\mathbf{b}_{h}\,\Vert\,\mathbf{m}\,]\bigr) \in [0,1]^{D},
\label{eq:bgtd_gate}
\end{equation}
where $W_{g}\!\in\!\mathbb{R}^{D\times 2D}$ and $\Vert$ denotes concatenation. The gated BH feature
\begin{equation}
\tilde{\mathbf{b}}_{h} \;=\; \mathbf{g}\odot\mathbf{b}_{h} \;+\; (\mathbf{1}-\mathbf{g})\odot\mathbf{m}
\label{eq:bgtd_fusion}
\end{equation}
is finally read out by a non-negative head $\hat{H}_{\mathrm{ave}} = \mathrm{ReLU}(W_{h}\tilde{\mathbf{b}}_{h})$. Equations~\eqref{eq:bgtd_gate}--\eqref{eq:bgtd_fusion} define the entire mechanism: the BH prediction is forced to use the shared representation as filtered through the convex combination of its own task-specific feature and the BF-derived morphology context, with the per-channel gate selecting how much of each to admit.

\paragraph{Design choices.} The choice of a sigmoid gate over a multi-head attention block or a softmax over keys is deliberate. With only one ``query'' source ($\mathbf{b}_{h}$) and one ``key/value'' source ($\mathbf{m}$), the multi-head formalism reduces to per-channel mixing while introducing parameters whose only purpose is to learn a 1-of-1 selection. A sigmoid gate retains the inductive bias --- adapt the BF-to-BH coupling strength to the local representation --- without paying that overhead, and lets each of the $D$ channels decide its own mixing weight, which we view as a soft per-feature analogue of the regime-conditional coupling discussed in Section~\ref{sec:prior}. The mechanism is also one-directional: morphology context flows from BF to BH but not back. This asymmetry reflects the underlying signal availability --- $\lambda_{p}$ is locally well-determined from optical and SAR signatures, while $H_{\mathrm{ave}}$ is poorly determined from non-VHR data and benefits from morphology hints --- and avoids the symmetric dual-gate variant whose extra capacity we found unnecessary in preliminary experiments. The full BGTD pathway adds approximately $1.7 \times 10^{4}$ trainable parameters on top of the encoder, less than 5\,\% of the model's total parameter count, so any accuracy gain it produces should be read as a contribution of \emph{wiring} rather than of capacity.

\subsection{MCL: Output-Level Cross-Task Coupling}
\label{sec:method:mcl}

\paragraph{Motivation.} BGTD couples the two tasks at the level of representations, but a model trained only with the standard two-head regression loss is still free to produce BH and BF predictions whose joint configuration drifts away from the FAR-consistent regime characterized in Section~\ref{sec:prior}. MCL closes this loop at the output level: it supervises a second, deliberately weak BH predictor that consumes only the footprint feature, and penalizes its disagreement with ground-truth BH. The result is gradient pressure on the BF branch to encode height-correlated structure --- not a hard FAR constraint, but a soft regulariser that prevents the cross-task representation from collapsing to a footprint-only summary.

\paragraph{Architecture.} A small auxiliary head reads height directly from the BF feature $\mathbf{b}_{f}\in\mathbb{R}^{D}$ defined in Section~\ref{sec:method:bgtd},
\[
\widehat{H}_{\mathrm{from\,BF}} \;=\; \mathrm{ReLU}\bigl(W_{2}\,\mathrm{GELU}(W_{1}\mathbf{b}_{f})\bigr),
\qquad W_{1}\!\in\!\mathbb{R}^{D/2\times D},\;W_{2}\!\in\!\mathbb{R}^{1\times D/2}.
\]
This head deliberately bypasses the BH branch and the cross-gate of BGTD, so $\widehat{H}_{\mathrm{from\,BF}}$ is the height implied by the BF lane alone. The non-negativity constraint mirrors the main BH head.

\paragraph{Training objective.} The full training loss combines task-specific Huber regressions, a Kendall--Gal uncertainty weighting between the two main tasks, and the consistency term:
\begin{align}
\mathcal{L}_{\mathrm{total}}
&= \tfrac{1}{2}e^{-\log\sigma_{h}^{2}}\,\mathcal{L}_{\mathrm{BH}}
   + \tfrac{1}{2}\log\sigma_{h}^{2} \nonumber \\
&\quad+ \tfrac{1}{2}e^{-\log\sigma_{f}^{2}}\,\mathcal{L}_{\mathrm{BF}}
       + \tfrac{1}{2}\log\sigma_{f}^{2}
       + \lambda_{c}(t)\,\mathcal{L}_{\mathrm{MCL}}, \label{eq:total_loss}
\end{align}
where $\mathcal{L}_{\mathrm{BH}} = \mathrm{Huber}_{\beta_{h}}(\hat{H}_{\mathrm{ave}},\,H^{*}_{\mathrm{ave}})$ and $\mathcal{L}_{\mathrm{BF}} = \mathrm{Huber}_{\beta_{f}}(\hat{\lambda}_{p},\,\lambda^{*}_{p})$ with task-appropriate Huber thresholds $\beta_{h}=2.0$\,m and $\beta_{f}=0.05$. The log-uncertainties $\log\sigma_{h}^{2}$ and $\log\sigma_{f}^{2}$ are learnable scalars that adaptively balance the two task losses across their different units. The morphology consistency term itself is
\begin{equation}
\mathcal{L}_{\mathrm{MCL}} \;=\; \mathrm{Huber}_{\beta_{h}}\bigl(\widehat{H}_{\mathrm{from\,BF}},\,H^{*}_{\mathrm{ave}}\bigr),
\end{equation}
i.e.\ the BF-derived BH surrogate is supervised against the same ground-truth height as the main BH head.

\paragraph{Design choices.} Three details deserve note. First, $\widehat{H}_{\mathrm{from\,BF}}$ is supervised by ground-truth height $H^{*}_{\mathrm{ave}}$, not by the main prediction $\hat{H}_{\mathrm{ave}}$: this prevents the surrogate from collapsing to a copy of the main head and forces the BF encoder to expose genuinely height-correlated structure rather than learn the main head's residual. Second, the consistency weight $\lambda_{c}$ follows a linear warmup over the first 10 epochs and saturates at 0.2; without warmup we observed the BF encoder being pulled toward height residuals before its own representation had stabilised, which destabilised both heads. Third, the Huber loss is preferred over MSE for BH (and consequently for MCL, since they share a target) because the right tail of the height distribution contains a thin population of high-rise cells whose squared residuals would otherwise dominate gradients; the Huber threshold $\beta_{h}=2.0$\,m is set to roughly the median absolute residual at convergence, which is the standard choice. The MCL pathway adds approximately $1.3 \times 10^{3}$ parameters (the consistency head only), so it inherits BGTD's ``wiring not capacity'' positioning: its empirical effect, reported in Section~\ref{sec:results}, must be read as a contribution of the loss term rather than of model capacity.

%% =============================================================
%% 5. Results
%% =============================================================
\section{Results}
\label{sec:results}

We organize the empirical evaluation around four hypothesis-driven questions, each one targeted at a specific claim made in the preceding sections.

\subsection{Q1: Does explicit cross-task coupling help, and by how much?}
\label{sec:results:q1}

We answer Q1 in two stages: an external comparison against three same-RF baselines that strip both proposed mechanisms (a Swin-MTL Transformer baseline plus ResNet-MTL and SENet-MTL CNN baselines), and an internal ablation that attributes the gap to BGTD and MCL individually. All metrics are reported on the held-out 54-city test split, on the checkpoint achieving the lowest validation combined MAE for each configuration. Numbers appear in Table~\ref{tab:q1_main}.

\begin{table*}[!htbp]
  \centering
  \footnotesize
  \caption{Test-split metrics for MorphoFormer and four same-RF baselines. Swin-MTL retains the single-stage Swin backbone of MorphoFormer with AMGE, MSMP, BGTD, and MCL disabled (shared encoder feeding two parallel regression heads); the two CNN baselines replace Swin with the corresponding convolutional encoder under the same input, split, dual-head, and uncertainty-weighted loss configuration. Best in bold.}
  \label{tab:q1_main}
  \begin{tabular}{lcccccccc}
    \toprule
    & \multicolumn{4}{c}{BH ($H_{\mathrm{ave}}$, m)} & \multicolumn{4}{c}{BF ($\lambda_{p}$)} \\
    \cmidrule(lr){2-5} \cmidrule(lr){6-9}
    Model & RMSE & MAE & CC & $R^{2}$ & RMSE & MAE & CC & $R^{2}$ \\
    \midrule
    Swin-MTL baseline   & 3.39 & 1.60 & 0.80 & 0.62 & 0.053 & 0.032 & 0.89 & 0.79 \\
    ResNet-MTL baseline & 3.40 & 1.60 & 0.79 & 0.62 & 0.058 & 0.036 & 0.87 & 0.75 \\
    SENet-MTL baseline  & 3.42 & 1.60 & 0.79 & 0.61 & 0.057 & 0.036 & 0.87 & 0.75 \\
    \midrule
    MorphoFormer (full) & \textbf{3.15} & \textbf{1.48} & \textbf{0.82} & \textbf{0.67} & \textbf{0.051} & \textbf{0.031} & \textbf{0.90} & \textbf{0.80} \\
    \quad w/o BGTD      & 3.26 & 1.54 & 0.81 & 0.65 & 0.051 & 0.032 & 0.90 & 0.80 \\
    \quad w/o MCL       & 3.25 & 1.53 & 0.81 & 0.65 & 0.051 & 0.031 & 0.90 & 0.81 \\
    \bottomrule
  \end{tabular}
\end{table*}

\begin{figure*}[!htbp]
  \centering
  \includegraphics[width=\textwidth]{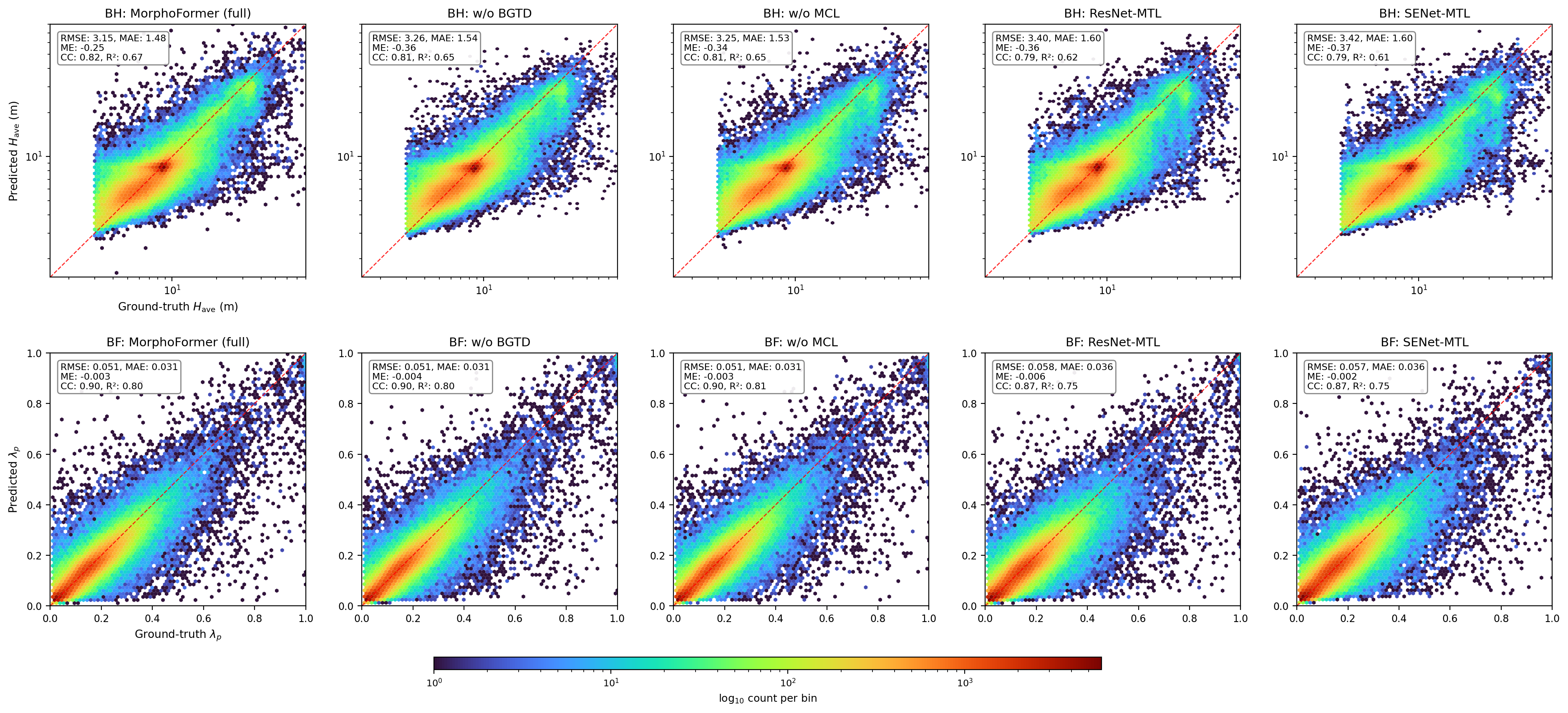}
  \caption{Predicted-vs-ground-truth hexbin densities on the test split for MorphoFormer (full and two ablations) and the ResNet-MTL / SENet-MTL CNN baselines at the same $9\times 9$ receptive field. Top row: BH (log-scaled axes). Bottom row: BF. The metric box in each panel reports RMSE / MAE (units: metres for BH, ratio for BF), ME, CC and $R^{2}$.}
  \label{fig:hexbin_compare}
\end{figure*}

A panel-of-hexbins comparison (Fig.~\ref{fig:hexbin_compare}) makes the predicted-vs-truth densities visually comparable across the three MorphoFormer configurations and the two CNN baselines. The diagonal mass tightens visibly when BGTD or MCL is restored, with an aggregate-metric improvement consistent with Table~\ref{tab:q1_main}; the ResNet-MTL and SENet-MTL panels show systematically wider scatter at the BH right tail and noticeable BF over-shrinkage.

\paragraph{External comparison.} MorphoFormer reduces BH test RMSE from 3.39\,m to 3.15\,m, a 0.24\,m (7.1\,\%) improvement over the Swin-MTL baseline at identical receptive field, modality set, and split; BH $R^{2}$ rises from 0.62 to 0.67, an 8.6\,\% relative uplift. BF $R^{2}$ improves marginally from 0.79 to 0.80, with BF already well-determined from optical and SAR signatures alone, so a ceiling effect is expected and the joint reduction is dominated by BH. The two CNN baselines (ResNet-MTL, SENet-MTL) cluster around the Swin-MTL baseline at 3.40\,m and 3.42\,m BH RMSE and 0.75 BF $R^{2}$, indicating that the gap to MorphoFormer is reproduced across encoder families and is therefore attributable to the cross-task wiring that all four baselines lack rather than to any specific backbone choice. Crucially, the BH RMSE gap exceeds by $\sim 3\times$ the $\approx 0.08$\,m ceiling that the marginal $(\lambda_{p}, H_{\mathrm{ave}})$ coupling sets in Section~\ref{sec:prior}. The model therefore clears the marginal-coupling ceiling --- consistent with the design of BGTD and MCL, which explicitly route BF-derived signal into the BH prediction rather than rely on the encoder to surface it.

\paragraph{Internal attribution.} The two BGTD/MCL ablations partition most of the 0.24\,m gap. Removing BGTD raises BH RMSE by 0.11\,m (3.15 $\to$ 3.26\,m) and drops BH $R^{2}$ from 0.67 to 0.65; removing MCL raises BH RMSE by 0.11\,m (3.15 $\to$ 3.25\,m) with the same $R^{2}$ drop. The two contributions are additive at this level of resolution, accounting jointly for 0.22\,m of the 0.24\,m external gap. BF $R^{2}$ is left within $\pm 0.01$ across both ablations, so the improvement is concentrated on the height task --- the same task that the marginal coupling analysis identified as carrying the most cross-task information available to recover.

\subsection{Q2: Where in the morphology plane does the coupling pay off?}
\label{sec:results:q2}

\begin{figure*}[!htbp]
  \centering
  \includegraphics[width=0.95\textwidth]{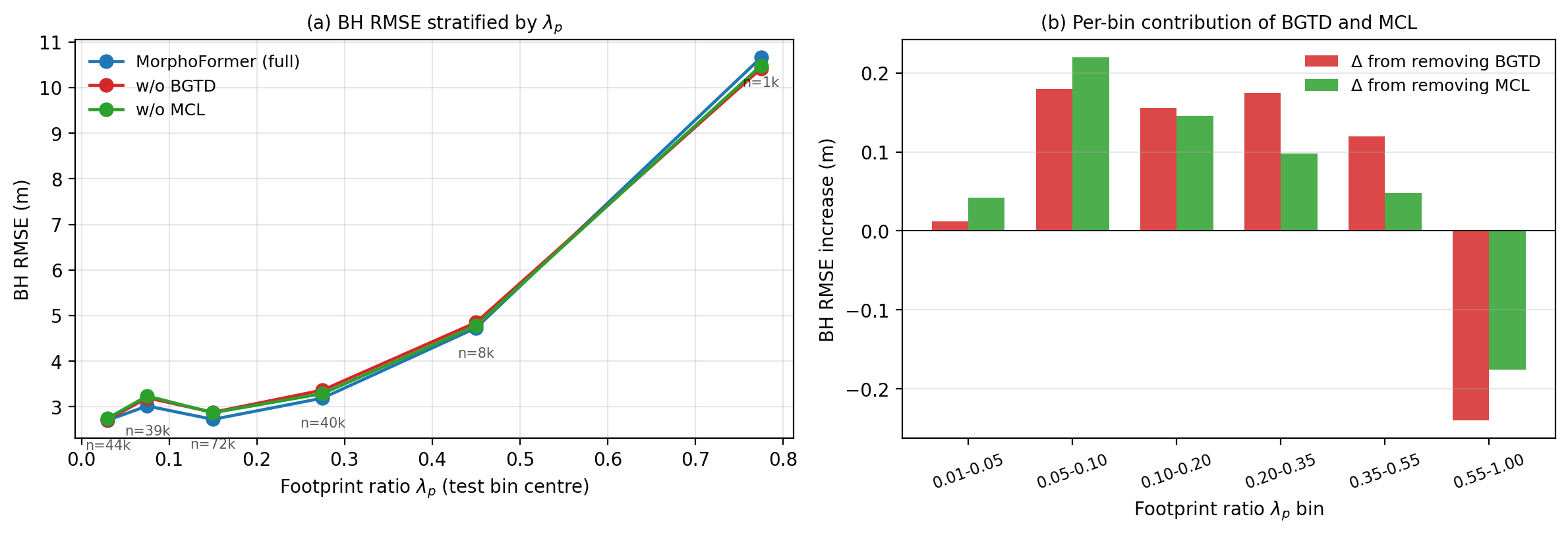}
  \caption{Stratification of test-set BH RMSE by $\lambda_{p}$ bin. (a) BH RMSE per bin for MorphoFormer (full) and the BGTD/MCL ablations; the dense-urban tail ($\lambda_{p}>0.55$) is dominated by a small population of high-rise cells with large absolute residuals. (b) Per-bin BH-RMSE increase upon removing each mechanism.}
  \label{fig:q2_strat}
\end{figure*}

The aggregate 0.24\,m gap of Q1 may be uniformly distributed across the test set or concentrated in particular morphological regimes. To distinguish the two we partition the 207\,579 test cells into six bins of ground-truth $\lambda_{p}$ and recompute BH RMSE per bin for the full model and the two ablations (Fig.~\ref{fig:q2_strat}; Table~\ref{tab:q2_strat}).

\begin{table}[!htbp]
  \centering
  \small
  \caption{Per-bin BH RMSE on the test split. ``$\Delta$'' columns give the increase upon removing each mechanism relative to the full model.}
  \label{tab:q2_strat}
  \begin{tabular}{lrcccc}
    \toprule
    $\lambda_{p}$ bin & $n$ & full & w/o BGTD ($\Delta$) & w/o MCL ($\Delta$) \\
    \midrule
    0.01--0.05 & 44\,425 & 2.704 & $+0.012$ & $+0.042$ \\
    0.05--0.10 & 39\,761 & 3.016 & $+0.180$ & $+0.219$ \\
    0.10--0.20 & 72\,419 & 2.723 & $+0.155$ & $+0.146$ \\
    0.20--0.35 & 40\,883 & 3.189 & $+0.175$ & $+0.098$ \\
    0.35--0.55 &  8\,056 & 4.728 & $+0.119$ & $+0.048$ \\
    0.55--1.00 &  1\,949 &10.657 & $-0.240$ & $-0.175$ \\
    \bottomrule
  \end{tabular}
\end{table}

Three patterns emerge. First, BGTD and MCL contribute most strongly in the mid-density regime $0.05 < \lambda_{p} < 0.35$, which is where the bulk of the test data lives (153\,k of the 207\,k cells, 74\,\%) and where the morphology context carries the most regression-relevant information: the conditional $H_{\mathrm{ave}}$ distributions in this regime are the ones that move most under conditioning on $\lambda_{p}$ (Fig.~\ref{fig:far_prior}b). The per-bin gain in this regime ranges from $+0.12$ to $+0.22$\,m for both ablations, considerably larger than the aggregate $0.11$\,m headline; the aggregate is averaged down by the sparser regimes where the gain is absent.

Second, the gain effectively vanishes for very low footprints ($\lambda_{p} < 0.05$, $n=44$\,k): removing BGTD changes BH RMSE by $+0.012$\,m, removing MCL by $+0.042$\,m. These cells are typically rural or peripheral with very few buildings, so the cross-task signal carries little marginal information --- there is little for the BF lane to encode that the shared encoder does not already capture from the bare presence/absence of structure in the bands. This negative result is itself informative: the mechanisms operate on cross-task structure that exists only where there is non-trivial built fabric to couple over, exactly as the prior characterisation predicts.

Third, the highest-density bin ($\lambda_{p} > 0.55$, $n=1\,949$) shows a small reversal: the ablations do slightly better than the full model. This regime contains $<1\,\%$ of the test cells --- a proportion that mirrors its share of the SHAFTS dataset overall, where dense high-rise morphology is sparsely sampled --- and BH RMSE in this bin is $\approx 10.7$\,m, four times the aggregate, because it is dominated by a thin tail of high-rise cells whose heights span tens of metres but whose footprint signatures are nearly saturated, leaving the BF channel essentially uninformative; in this regime the cross-task pathway provides no usable signal and the gain mechanisms cannot generalise into it. The reversal therefore reflects the limit of the prior, not a contradiction of it.

Taken together, the stratification confirms the more general claim of Section~\ref{sec:prior}: the BGTD and MCL mechanisms extract cross-task signal precisely where the prior says it exists, neither earlier nor later. The aggregate $0.24$\,m gap of Q1 is not uniformly distributed; it is concentrated in the morphological regime that the prior characterisation already identified as the one carrying useful cross-task structure.

\subsection{Q3: How does the BGTD gate operate at convergence?}
\label{sec:results:q3}

\begin{figure*}[!htbp]
  \centering
  \includegraphics[width=\textwidth]{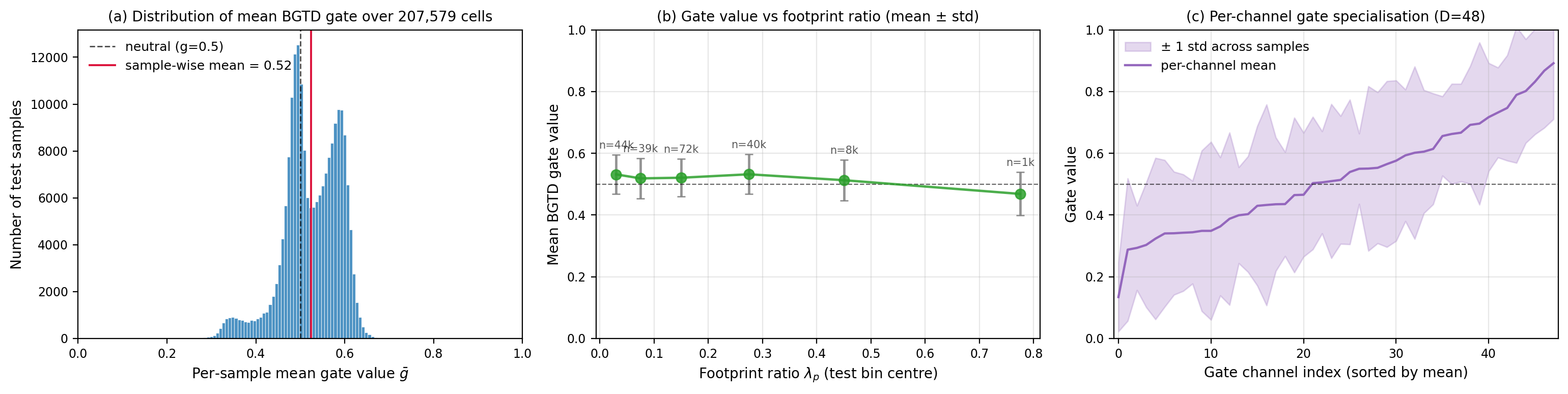}
  \caption{BGTD cross-gate activation on the test split. (a) Per-sample mean gate value $\bar{g}$ over 207\,579 cells. (b) $\bar{g}$ stratified by $\lambda_{p}$ bin (mean $\pm$ std). (c) Per-channel mean and 1-std band across samples, with channels sorted by mean.}
  \label{fig:q3_gate}
\end{figure*}

The Q1--Q2 results establish that BGTD is responsible for roughly half of the headline accuracy gain. This subsection looks inside the trained gate: we extract the cross-gate output $\mathbf{g}\in[0,1]^{D}$ from Eq.~\eqref{eq:bgtd_gate} on every test cell and characterise its behaviour at three granularities (Fig.~\ref{fig:q3_gate}).

Three observations follow.

First, the per-sample mean $\bar{g}$ has a sample-wise mean of 0.524 with standard deviation 0.064 (Fig.~\ref{fig:q3_gate}a). The distribution is visibly bimodal, with peaks near 0.48 and 0.58, but the spread between them is small ($<\!0.10$ in $\bar{g}$): in aggregate the gate operates close to neutral mixing of $\mathbf{b}_{h}$ and $\mathbf{m}$. This argues against an interpretation in which the gate switches sharply between ``use BF context'' and ``use BH feature'' regimes on a per-input basis.

Second, the per-bin stratification is essentially flat (Fig.~\ref{fig:q3_gate}b). Across the five bins that contain $>\!8\,000$ cells the per-bin gate mean lies in $[0.51, 0.53]$, only the highest-density bin ($\lambda_{p}>0.55$, $n=1\,949$) drops modestly to $0.47$ as the model leans slightly more on the morphology context for tall, dense cores. The gate is therefore not the mechanism through which BGTD's regime-specific behaviour identified in Q2 is realised: per-input adaptation explains very little of the per-bin gain seen in Fig.~\ref{fig:q2_strat}.

Third, the per-channel statistics tell a different story (Fig.~\ref{fig:q3_gate}c). Sorting the $D=48$ channels by their mean gate value reveals a smooth ramp from $\approx 0.13$ to $\approx 0.89$. Roughly a fifth of the channels operate as ``BF-dominant'' channels with $\bar{g}_{c}<0.3$ across most samples (so $\tilde{\mathbf{b}}_{h,c}\approx\mathbf{m}_{c}$), another fifth are ``BH-dominant'' with $\bar{g}_{c}>0.7$, and the remaining channels mix the two more evenly. The per-channel standard deviation across samples is moderate (median $0.18$), so the channel specialisation is largely fixed at convergence rather than dynamically reshuffled per input.

The three observations together suggest that BGTD's contribution should be understood as a \emph{learned, per-channel routing} between BH-derived and BF-derived feature subspaces, rather than as an input-conditional switch. This is consistent with the design rationale in Section~\ref{sec:method:bgtd}: the sigmoid gate was chosen precisely to permit per-channel mixing without imposing an input-conditional inductive bias, and the converged behaviour validates that choice. It also clarifies a result of Q1: removing BGTD does not just remove a small adaptive correction --- it removes the model's mechanism for routing approximately 40\,\% of the BH feature dimension through the BF lane, which is an architecturally substantial change explaining the 0.11\,m aggregate impact.

\subsection{Q4: Does the height-from-footprint surrogate behave like a height predictor?}
\label{sec:results:q4}

\begin{figure*}[!htbp]
  \centering
  \includegraphics[width=0.95\textwidth]{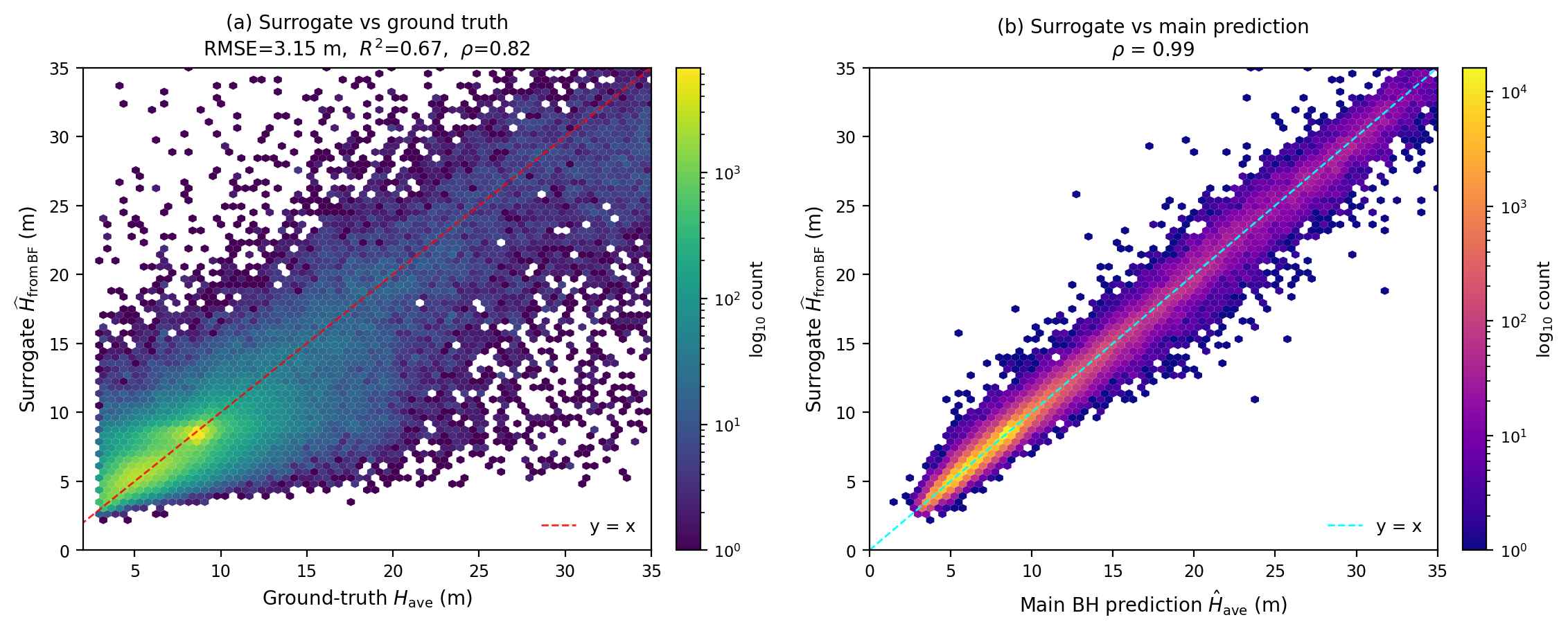}
  \caption{Behaviour of the height-from-footprint surrogate on the test split. (a) Hexbin scatter of ground-truth $H_{\mathrm{ave}}$ against the surrogate $\widehat{H}_{\mathrm{from\,BF}}$. (b) Hexbin scatter of the main BH prediction $\hat{H}_{\mathrm{ave}}$ against the surrogate.}
  \label{fig:q4_surr}
\end{figure*}

The MCL term of Eq.~\eqref{eq:total_loss} works only if the auxiliary head is genuinely doing what its name implies, namely predicting BH from a BF-derived feature. We therefore ask whether the surrogate $\widehat{H}_{\mathrm{from\,BF}}$ behaves as an independent height predictor at convergence (Fig.~\ref{fig:q4_surr}, Table~\ref{tab:q4_surr}).

\begin{table}[!htbp]
  \centering
  \small
  \caption{Test-set behaviour of the BH-from-BF surrogate vs the main BH head on the same test cells.}
  \label{tab:q4_surr}
  \begin{tabular}{lcccc}
    \toprule
    Predictor & RMSE (m) & MAE (m) & $R^{2}$ & $\rho$(\,$\cdot$\,, GT) \\
    \midrule
    Main BH head $\hat{H}_{\mathrm{ave}}$              & 3.145 & 1.483 & 0.671 & 0.820 \\
    Surrogate $\widehat{H}_{\mathrm{from\,BF}}$        & 3.154 & 1.510 & 0.669 & 0.819 \\
    \bottomrule
  \end{tabular}
\end{table}

The surrogate's accuracy on ground-truth BH is, to two decimal places, indistinguishable from the main head (RMSE 3.15 vs 3.15, $R^{2}$ 0.67 vs 0.67); it differs by only 0.01\,m in RMSE and 0.002 in $R^{2}$. This is a strong outcome for MCL: the auxiliary head, which sees only the footprint feature $\mathbf{b}_{f}$ and bypasses the BGTD cross-gate entirely, nonetheless recovers BH almost as well as the main head --- meaning the consistency loss has pushed the BF feature subspace to encode essentially the same height information that the full BH lane carries.

The relationship between the surrogate and the main prediction (Fig.~\ref{fig:q4_surr}b) sharpens this picture. The two predictors are correlated at $\rho = 0.991$, far above either one's correlation with ground truth ($\rho = 0.82$). The two pathways therefore agree with each other much more tightly than either agrees with the truth: the residual disagreement between $\hat{H}_{\mathrm{ave}}$ and $\widehat{H}_{\mathrm{from\,BF}}$ is small relative to the residual disagreement of either with $H^{*}_{\mathrm{ave}}$. We read this as evidence that MCL has succeeded as a soft FAR-consistency regulariser in the precise form intended in Section~\ref{sec:method:mcl}: under a converged training, the BF lane has been forced to encode height-correlated structure to such a degree that a small two-layer probe over $\mathbf{b}_{f}$ can recover BH almost as well as the main, BGTD-coupled BH head can.

%% =============================================================
%% 6. Discussion
%% =============================================================
\section{Discussion}
\label{sec:discussion}

\subsection{What the empirical evidence supports}

The paper opened with two propositions: that the marginal $(\mathrm{BH},\,\mathrm{BF})$ coupling sets a tight, data-driven ceiling on what any shared-encoder predictor can extract, and that exceeding this ceiling requires explicit cross-task routing rather than reliance on the encoder. The four results sections together support a tightened version of these claims. The empirical ceiling implied by the marginal coupling is $\approx 0.08$\,m of BH RMSE reduction (Section~\ref{sec:prior}); MorphoFormer reduces BH RMSE by 0.24\,m relative to the Swin-MTL baseline at the same receptive field (Section~\ref{sec:results:q1}), about three times that ceiling. The bulk of this 0.24\,m gap is accounted for by the two proposed mechanisms (BGTD: 0.11\,m, MCL: 0.11\,m). The model therefore extracts cross-task structure that exceeds the marginal-coupling ceiling; explicit routing of BF-derived signal into the BH prediction is the channel through which it does so.

The Q2--Q4 analyses sharpen the interpretation in three ways. Q2 shows that the gain is regime-specific: it concentrates in the mid-density regime $0.05<\lambda_{p}<0.35$ where the bulk of the data lives and where conditional $H_{\mathrm{ave}}$ distributions move most under conditioning on $\lambda_{p}$ (Fig.~\ref{fig:far_prior}b), and vanishes in the very-low-footprint regime where there is little built fabric over which to couple. Q3 demonstrates that the BGTD gate operates by per-channel specialisation, with about $40\,\%$ of the 48 hidden channels routed dominantly through the BF lane (Fig.~\ref{fig:q3_gate}c); this is structurally a substantial change to the BH branch, not a small adaptive correction, and it explains why removing BGTD costs as much as it does. Q4 shows that MCL has succeeded as a soft FAR-consistency regulariser: a two-layer probe over $\mathbf{b}_{f}$ recovers BH at $R^{2}=0.67$, indistinguishable from the main head's $R^{2}=0.67$, and the surrogate-main agreement of $\rho=0.991$ exceeds either one's agreement with ground truth ($\rho\approx 0.82$).

\subsection{Where the gain ends: the high-rise dense regime}

\begin{figure*}[!htbp]
  \centering
  \includegraphics[width=\textwidth]{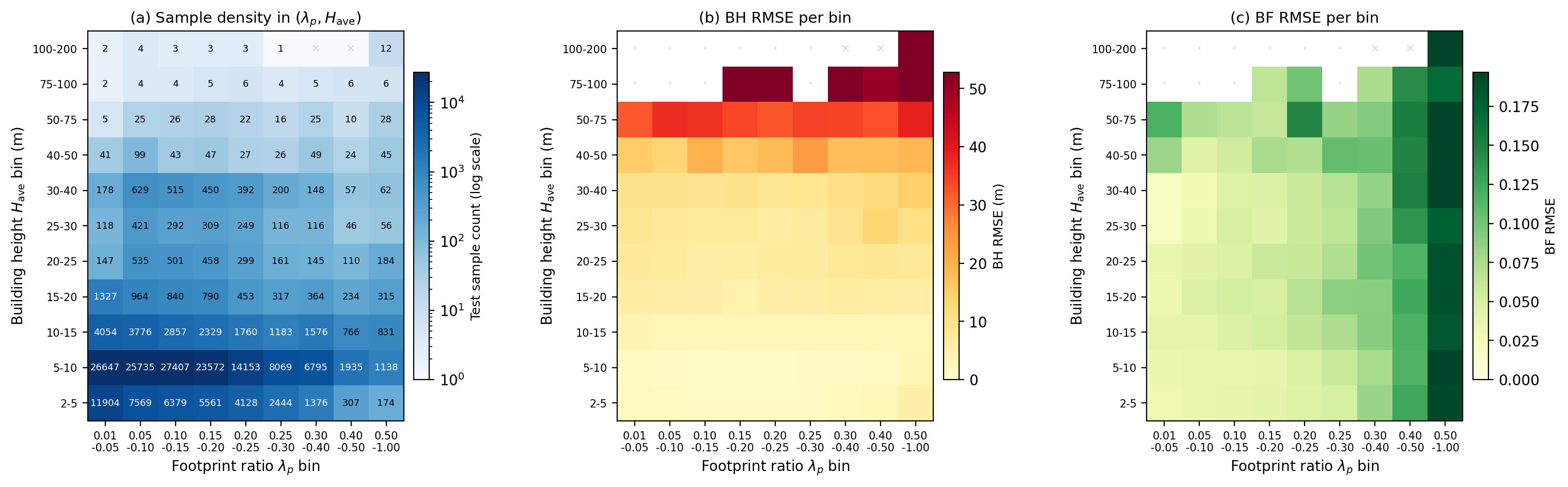}
  \caption{Error origins for MorphoFormer on the test split, binned over $(\lambda_{p}, H_{\mathrm{ave}})$. (a) Sample count per bin (log scale, with raw counts annotated). (b) BH RMSE per bin (m). (c) BF RMSE per bin. Bins with $<\!5$ samples are blanked. High-error regions in (b) and (c) coincide with the sparsely-populated bins in (a), and these sparse bins reflect the SHAFTS dataset's underrepresentation of dense high-rise morphology rather than a property of the test split.}
  \label{fig:err_source}
\end{figure*}

The high-density tail $\lambda_{p}>0.55$ shows a small reversal in Q2 in which the ablations marginally outperform the full model. The bin contains under 1\,\% of the test cells, and BH RMSE in it is dominated by a thin population of high-rise cells whose footprint signatures are nearly saturated and whose heights span tens of metres; in this regime the BF channel itself is essentially uninformative, so the cross-task pathway has nothing to add. The two-dimensional binning of Fig.~\ref{fig:err_source} sharpens this picture: large absolute errors concentrate in the upper-right corner of the $(\lambda_{p}, H_{\mathrm{ave}})$ plane, with only 131 of 207\,579 test cells falling in the $H_{\mathrm{ave}}>30$\,m and $\lambda_{p}>0.30$ region. This sparsity is a property of the SHAFTS dataset itself, not of our test split: the same morphological corner contains only $\sim 7{,}600$ of the 1.88\,M filtered training cells (about 0.4\,\%). The high errors at extreme cells therefore reflect the upstream dataset's sampling of dense high-rise morphology rather than the architecture, and are expected to attenuate as denser high-rise reference sources become available.

\subsection{Implications for cross-task design in remote-sensing regression}

The headline takeaway is that joint BH/BF retrieval benefits substantially from explicit cross-task wiring: BGTD and MCL together drove BH RMSE down by 0.22\,m on this dataset --- about three times the $\approx 0.08$\,m ceiling that the marginal $(\lambda_{p}, H_{\mathrm{ave}})$ coupling allows in principle, and a contribution of the same order as the engineered priors that joint regression in remote sensing typically relies on shared encoders to discover. Mechanisms that route BF-derived signal into the BH prediction are doing work that no shared encoder is required to do, and that work is large.

A complementary methodological takeaway concerns the role of empirical prior characterisation. By reporting the marginal $(\lambda_{p}, H_{\mathrm{ave}})$ coupling and its 0.08\,m RMSE-reduction ceiling \emph{before} introducing any architectural choice, we anchor the architectural contribution to a data-driven benchmark rather than to an author-elected baseline. The 0.16\,m of improvement beyond the marginal ceiling is therefore directly interpretable as the contribution of explicit cross-task routing, and we propose this prior-first reporting as a transferable practice for joint-regression problems in remote sensing where the targets are physically coupled.

A second complementary observation concerns the role of input context. Any cross-task routing mechanism is only as informative as the morphology that the encoder is allowed to see: the 9$\times$9 / 900\,m receptive field used here is what makes neighbourhood-level FAR structure visible in the first place, and we expect cross-task gains of this kind to scale with the morphological scope the input window admits.

\subsection{Outlook}

The 54-city training and test set spans a wide range of urban morphology and density, and supports the architectural claims of this paper. Extending the same cross-task mechanism to additional regions and to finer-resolution sources is a natural application of the design: because BGTD and MCL operate on cross-task representations rather than on pixels, the routing principle they implement carries directly to 10\,m Sentinel-2 ARD or sub-meter VHR settings, and finer morphology signals are likely to amplify the same coupling. The BF $\to$ BH directionality of the two mechanisms reflects the asymmetry of signal availability in non-VHR sources --- $\lambda_{p}$ is locally well-determined while $H_{\mathrm{ave}}$ is not --- and is the configuration we use throughout.

%% =============================================================
%% 7. Conclusion
%% =============================================================
\section{Conclusion}
\label{sec:conclusion}

We have argued that the most under-exploited direction for improving joint estimation of building height and footprint is the explicit encoding of the (BH,\,BF) coupling implied by floor-area-ratio physics. An empirical characterisation on 1.88\,M training cells anchors this argument quantitatively: $\lambda_{p}$ alone accounts for 4.6\,\% of the variance in $H_{\mathrm{ave}}$, which is a tight, data-driven ceiling of about 0.08\,m on the BH RMSE reduction extractable from the marginal coupling, and the threshold any explicit cross-task mechanism must clear to demonstrate value beyond what a shared-encoder formulation already passively offers. We instantiated such a mechanism as MorphoFormer: a BF-Guided Task Decoder that routes a footprint-derived morphology context into the height branch via a per-channel sigmoid gate, and a Morphology Consistency Loss that supervises an auxiliary height-from-footprint surrogate against ground-truth height. On a strict 54-city geo-blocked split, MorphoFormer reduces BH test RMSE from 3.39 to 3.15\,m relative to a Swin-MTL baseline at the same receptive field; controlled ablations attribute the bulk of the 0.24\,m gap to BGTD ($0.11$\,m) and MCL ($0.11$\,m), exceeding the marginal-coupling ceiling by approximately $3\times$. The gain concentrates in the morphological regime that the prior characterisation identified as most informative, and the auxiliary head behaves like a competent independent height predictor at convergence. Because both mechanisms operate on cross-task representations rather than on pixels, the design carries no intrinsic dependence on input resolution and the same routing principle is expected to apply directly to 10\,m or sub-meter VHR settings.

%% ---- Acknowledgements ----
\section*{Acknowledgements}
This research was supported by the 2023-MOIS36-004 (RS-2023-00248092) of the Technology Development Program on Disaster Restoration Capacity Building and Strengthening, funded by the Ministry of Interior and Safety (MOIS, Republic of Korea).

%% ---- Declaration of competing interest ----
\section*{Declaration of competing interest}
The authors declare that they have no known competing financial interests or personal relationships that could have appeared to influence the work reported in this paper.

%% ---- Data and code availability ----
\section*{Data and code availability}
The SHAFTS reference dataset used for training and evaluation is openly available at \url{https://zenodo.org/records/6370003} \cite{shafts2023}. Sentinel-1 GRD Level-1, Sentinel-2 Level-2A, and SRTM V3 DEM imagery are publicly accessible through the Google Earth Engine catalogue. The MorphoFormer source code, trained model weights, and prediction artefacts will be released at \url{https://github.com/geumjin99/morphoformer} upon acceptance of the manuscript.

%% ---- Bibliography ----
\bibliographystyle{elsarticle-num}
\bibliography{cas-refs,new_refs}

\end{document}